# Closed-Form Learning of Markov Networks from Dependency Networks

**Daniel Lowd**
Dept. of Computer and Information Science
University of Oregon
Eugene, OR 97403
lowd@cs.uoregon.edu

## Abstract

Markov networks (MNs) are a powerful way to compactly represent a joint probability distribution, but most MN structure learning methods are very slow, due to the high cost of evaluating candidates structures. Dependency networks (DNs) represent a probability distribution as a set of conditional probability distributions. DNs are very fast to learn, but the conditional distributions may be inconsistent with each other and few inference algorithms support DNs. In this paper, we present a closed-form method for converting a DN into an MN, allowing us to enjoy both the efficiency of DN learning and the convenience of the MN representation. When the DN is consistent, this conversion is exact. For inconsistent DNs, we present averaging methods that significantly improve the approximation. In experiments on 12 standard datasets, our methods are orders of magnitude faster than and often more accurate than combining conditional distributions using weight learning.

## 1 INTRODUCTION

Joint probability distributions are useful for representing and reasoning about uncertainty in many domains, from robotics to medicine to molecular biology. One of the most powerful and popular ways to represent a joint probability distribution compactly is with a Markov network (MN), an undirected probabilistic graphical model. Inference in an MN can compute marginal or conditional probabilities, or find the most probable configuration of a subset of variables given evidence. Although exact inference is usually intractable, numerous approximate inference algorithms exist to exploit the structure present in the MN representation.

One of the largest disadvantages of Markov networks is the difficulty of learning them from data. In the fully observed case, weight learning is a convex optimization problem, but cannot be done in closed form except in very special circumstances, e.g. [18]. Structure learning is even harder, since it typically involves a search over a large number of candidate structures, and weights must be learned for each candidate before it can be scored [4, 14, 3]. An increasingly popular alternative is to use local methods to select the structure, by finding the dependencies for each variable separately and combining them using weight learning [15, 12].

Dependency networks (DNs) [8] represent the joint distribution itself as a set of conditional probability distributions, one for each variable. In terms of representational power, DNs are comparable to MNs, since every MN can be represented as a consistent DN and vice versa. The advantage of DNs is that they are much easier to learn than MNs, since each conditional probability distribution can be learned separately. However, the local distributions may not be consistent with any joint distribution, making the model difficult to interpret. Furthermore, very few inference algorithms support DNs. As a result, DNs remain much less popular than MNs.

In this paper, we present a new method for converting dependency networks into Markov networks, allowing us to enjoy both the efficiency of DN learning and the convenience of the MN representation. Surprisingly, this translation can be done in closed-form without any kind of search or optimization. If the DN is consistent, then this conversion is exact. For the general case of an inconsistent DN, we present averaging methods that significantly improve the approximation. As long as the maximum feature length is bounded by a constant, our method runs in linear time and space with respect to the size of the original DN.

Our method is similar in spirit to previous work on learning MNs by combining local, conditional distributions with weight learning [12, 15]. However, instead of ignoring the parameters of those local distributions, we use them to directly compute the parameters of the joint distribution. Hulten et al. [9] proposed a method for converting DNs to Bayesian networks. However, this conversion process re-quired searching through possible structures and led to less accurate models than learning a BN directly from data.

In experiments on 12 standard datasets, we find that our DN2MN conversion method is orders of magnitude faster than weight learning and often more accurate. For DNs with decision tree conditional distributions, the converted MN was often more accurate than the original DN. With logistic regression conditional distributions, the converted MNs were significantly more accurate than performing weight learning.

Our method has potential applications outside of MN structure learning as well. For domains where data is limited or unavailable, an expert could specify the conditional distributions of a DN, which could then be translated into the joint distribution of an MN. This could be much easier and less error-prone than attempting to specify the entire joint distribution directly.

Our paper is organized as follows. We begin with background on Markov networks and dependency networks in Sections 2 and 3. In Section 4, we describe how to convert consistent dependency networks into Markov networks that represent the exact same probability distribution. In Section 5, we discuss inconsistent conditional probability distributions and methods for improving the approximation quality through averaging. We evaluate our methods empirically in Section 6 and conclude in Section 7.

## 2 MARKOV NETWORKS

A *Markov network* (MN) represents a probability distribution over a set of random variables $\mathbf{X} = \{X_1, X_2, \ldots, X_n\}$ as a normalized product of factors:

$$P(\mathbf{X}) = \frac{1}{Z} \prod_i \phi_i(\mathbf{D}_i) \quad (1)$$

$Z$ is the partition function, a normalization constant to make the distribution sum to one; $\phi_i$ is the $i$th factor, sometimes referred to as a potential function; and $\mathbf{D}_i \subset \mathbf{X}$ is the set of variables in the domain of $\phi_i$.

A probability distribution where $P(\mathbf{x}) > 0$ for all $\mathbf{x} \in \mathbf{X}$ is said to be *positive*. An MN that represents a positive distribution can also be written as a *log-linear model*. In a log-linear model, the probability distribution is expressed as an exponentiated weighted sum of feature functions $f_i(\mathbf{D}_i)$ rather than as a product of factors:

$$P(\mathbf{X}) = \frac{1}{Z} \exp\left(\sum_i w_i f_i(\mathbf{D}_i)\right) \quad (2)$$

The correspondence between (1) and (2) is easily shown by letting $f_i = \log \phi_i$ and $w_i = 1$.

For discrete domains, a common choice is to use conjunctions of variable tests as the features. Each test is of the form $(X_i = v_i)$, where $X_i$ is a variable in $\mathbf{X}$ and $v_i$ is a value of that variable. We sometimes abbreviate tests of Boolean variables, so that $(X_i = T)$ is written as $X_i$ and $(X_i = F)$ is written as $\neg X_i$. A conjunctive feature equals 1 if its arguments satisfy the conjunction and 0 otherwise. Any factor represented as a table can be converted into a set of conjunctive features with one feature for each entry in the table. When the factors have repeated values or other types of structure, a feature-based representation is often more compact than a tabular one.

The *Markov blanket* of a variable $X_i$, denoted $\text{MB}(X_i)$, is the set of variables that render $X_i$ independent from all other variables in the domain. In an MN, this set consists of all variables that appear in a factor or feature with $X_i$. These independencies, and others, are entailed by the factorization in (1).

### 2.1 INFERENCE

Given an MN, we often wish to answer queries such as the probability of one or more variables given evidence. In general, computing exact marginal and conditional probabilities is #P-complete [16], so approximate inference algorithms are commonly used instead. One of the most popular is Gibbs sampling [5].

Gibbs sampling is a Markov chain Monte Carlo method that generates a set of random samples and uses them to answer queries. The sampler is initialized to a random state consistent with the evidence. Samples are generated by resampling each non-evidence variable in turn, according to its conditional probability given the current states of the other variables. In early samples, the sampler may be in an unlikely and non-representative state, so these "burn-in" samples are typically excluded from consideration. The probability of a query is estimated as the fraction of samples consistent with the query. For positive distributions, Gibbs sampling will eventually converge to the correct probabilities, but this can take a very long time and convergence can be difficult to detect.

### 2.2 WEIGHT LEARNING

In maximum likelihood parameter estimation, the goal is to select a set of parameters that maximizes the log-likelihood of the model on the training data. Here we assume that the MN is expressed as a log-linear model with a fixed set of features and that all variables are observed in the training data. Since the log-likelihood of an MN is a concave function of the weights, parameter learning can be framed as a convex optimization problem and solved with standard gradient-based approaches. Since log-likelihood and its gradient are usually intractable to compute exactly and slow to approximate, a commonly-used alternative is pseudo-likelihood. Pseudo-log-likelihood (PLL) [2] is the sum of the conditional log-likelihood of each variable given

the other variables:

$$\log P_w^\bullet(\mathbf{X}=\mathbf{x}) = \sum_{i=1}^{n} \log P_w(X_i=x_i|\mathbf{X}_{-i}=\mathbf{x}_{-i})$$

where $\mathbf{X}_{-i} = \mathbf{X} - X_i$, the set of all variables except for $X_i$. Unlike log-likelihood, PLL and its gradient can both be evaluated efficiently. Like log-likelihood, PLL is a concave function, so any local optimum is also a global optimum. The main disadvantage of PLL is that it tends to handle long-range dependencies poorly. Optimizing PLL optimizes the model's ability to predict individual variables given large amounts of evidence, and this may lead to worse performance when less evidence is available. To control overfitting, a zero-mean Gaussian prior is typically placed on each weight.

### 2.3 STRUCTURE LEARNING

The goal of MN structure learning is to find a set of factors and parameters or features and weights with a high score on the training data. As with weight learning, pseudo-likelihood is a common choice due to the intractability of the partition function. Of the many MN structure learning variations that have been proposed, we touch on just a few algorithms and their main themes.

One common approach to structure learning is to perform a greedy search through the space of possible structures. Della Pietra et al. [4] conduct this search in a top-down manner, starting with atomic features (individual variable tests) and extending and combining these simple features until convergence. Davis and Domingos [3] propose a bottom-up search instead, creating many initial features based on the training data and repeatedly merging features as learning progresses. Search-based algorithms tend to be slow, since scoring candidate structures can be expensive and there are many candidate structures to consider.

Another approach is to use local models to find the Markov blanket of each variable separately and then combine the features of these local models into a global structure. Ravikumar et al. [15] do this by learning a logistic regression model with L1 regularization for each variable. The use of L1 regularization encourages sparsity, so that most of the interaction weights are zero. All non-zero interactions are turned into conjunctive features. In other words, if the logistic regression model for predicting $X_i$ has a non-zero weight for $X_j$ then the feature $X_i \wedge X_j$ is added to the model. Lowd and Davis [12] adopt a similar approach, but use probabilistic decision trees as the local model. Each decision tree is converted into a set of conjunctive features by considering each path from the tree root to a leaf as a conjunction of variable tests. In both works, the authors select feature weights by performing weight learning. In addition to being much faster than the search-based methods, the local approaches often lead to more accurate models.

## 3 DEPENDENCY NETWORKS

A *dependency network* (DN) [8] consists of a set of conditional probability distributions (CPDs) $P_i(X_i|\text{MB}(X_i))$, each defining the probability of a single variable given its Markov blanket. A DN is said to be *consistent* if there exists a probability distribution $P$ that is consistent with the DN's conditional distributions. Inconsistent DNs are sometimes called *general* dependency networks.

Since Gibbs sampling only uses the conditional probability of each variable given its Markov blanket, it is easily applied to DNs. The probability distribution represented by a DN is defined as the stationary distribution of the Gibbs sampler, given a fixed variable order. If the DN is consistent, then its conditional distributions must be consistent with some MN, and Gibbs sampling converges to the same distribution as the MN. If the DN is inconsistent, then the stationary distribution may depend on the order in which the variables are resampled. Furthermore, the joint distribution determined by the Gibbs sampler may be inconsistent with the conditional probabilities in the CPDs.

Few other inference algorithms have been defined for DNs. Heckerman et al. [8] describe a method for using Gibbs sampling and CPD values together to estimate rare probabilities with lower variance; Toutanova et al. [17] do maximum a posteriori (MAP) inference in a chain-structured DN with an adaptation of the Viterbi algorithm; and Lowd and Shamaei [13] propose mean field inference for DNs. We know of no other methods.

A DN can be constructed from any MN by constructing a conditional probability distribution for each variable given its Markov blanket. A DN can also be learned from data by learning a probabilistic classifier for each variable. This makes DN learning trivial to parallelize with a separate process for each variable. However, when learned from data, the resulting CPDs may be inconsistent with each other.

Any standard method for learning a probabilistic classifier can be used. Heckerman et al. [8] learn a probabilistic decision tree for each variable. In a probabilistic decision tree, the interior nodes of the tree are variable tests (such as $(X_4 = T)$), the branches are labeled with test outcomes (such as *true* and *false*), and the leaves specify the marginal distribution of the target variable given the tests on the path from the root to leaf. Probabilistic decision trees can be learned greedily to maximize the conditional log-likelihood of the target variable.

Hulten et al. [9] investigated the problem of converting a DN to a Bayesian network (BN). However, rather than trying to match the distribution as closely as possible, they tried to find the highest-scoring acyclic subset of the DN structure. After proving the problem to be NP-hard, they proposed a greedy algorithm for removing the least essential edges. The resulting BN was consistently less accurate than training one from scratch. Furthermore, it only applies to DNs with tree CPDs.

## 4 CONVERTING CONSISTENT DEPENDENCY NETWORKS

We now discuss how to construct a joint distribution from a set of positive conditional distributions. To begin with, we assume that the conditional distributions, $P(X_i|\mathbf{X}_{-i})$, are the conditionals of some unknown joint probability distribution $P$. Later, we will relax this assumption and discuss how to find the most effective approximation.

Consider two instances, $\mathbf{x}$ and $\mathbf{x}'$, that only differ in the state of one variable, $X_j$. In other words, $x_i = x_i'$ for all $i \neq j$, so $\mathbf{x}_{-j} = \mathbf{x}'_{-j}$. We can express the ratio of their probabilities as follows:

$$\frac{P(\mathbf{x})}{P(\mathbf{x}')} = \frac{P(x_j, \mathbf{x}_{-j})}{P(x_j', \mathbf{x}_{-j})} = \frac{P(x_j|\mathbf{x}_{-j})P(\mathbf{x}_{-j})}{P(x_j'|\mathbf{x}_{-j})P(\mathbf{x}_{-j})} = \frac{P(x_j|\mathbf{x}_{-j})}{P(x_j'|\mathbf{x}_{-j})} \quad (3)$$

Note that the final expression only involves a conditional distribution, $P(X_j|X_{-j})$, not the full joint. The conditional distribution must be positive in order to avoid division by zero.

If $\mathbf{x}$ and $\mathbf{x}'$ differ in multiple variables, then we can express their probability ratio as the product of multiple single-variable transitions. We construct a sequence of instances $\{\mathbf{x}^{(0)}, \mathbf{x}^{(1)}, \ldots, \mathbf{x}^{(n)}\}$, each instance differing in at most one variable from the previous instance in the sequence. Let order $o \in S_n$ be a permutation of the numbers 1 to $n$ and let $o[i]$ refer to the $i$th number in the order. We define $\mathbf{x}^{(i)}$ inductively:

$$\mathbf{x}^{(0)} = \mathbf{x}$$
$$\mathbf{x}^{(i)} = (x'_{o[i]}, \mathbf{x}^{(i-1)}_{-o[i]})$$

In other words, the $i$th instance $\mathbf{x}^{(i)}$ simply changes the $o[i]$th variable from $x_{o[i]}$ to $x'_{o[i]}$ and is otherwise identical to the previous element, $\mathbf{x}^{(i-1)}$. Thus, in $\mathbf{x}^{(i)}$ the first $i$ variables in the order are set to their values in $\mathbf{x}'$ and the latter $n-i$ are set to their values in $\mathbf{x}$. Note that $\mathbf{x}^{(n)} = \mathbf{x}'$, since all $n$ variables have been changed to their values in $\mathbf{x}'$.

We can use these intermediate instances to express the ratio $P(\mathbf{x})/P(\mathbf{x}')$ as a product:

$$\frac{P(\mathbf{x})}{P(\mathbf{x}')} = \frac{P(\mathbf{x}^{(0)})}{P(\mathbf{x}^{(n)})}$$
$$= \frac{P(\mathbf{x}^{(0)})}{P(\mathbf{x}^{(n)})} \times \frac{P(\mathbf{x}^{(1)})}{P(\mathbf{x}^{(1)})} \times \cdots \times \frac{P(\mathbf{x}^{(n-1)})}{P(\mathbf{x}^{(n-1)})}$$
$$= \frac{P(\mathbf{x}^{(0)})}{P(\mathbf{x}^{(1)})} \times \frac{P(\mathbf{x}^{(1)})}{P(\mathbf{x}^{(2)})} \times \cdots \times \frac{P(\mathbf{x}^{(n-1)})}{P(\mathbf{x}^{(n)})}$$
$$= \prod_{i=1}^{n} \frac{P(\mathbf{x}^{(i-1)})}{P(\mathbf{x}^{(i)})} = \prod_{i=1}^{n} \frac{P(x_{o[i]}|\mathbf{x}^{(i)}_{-o[i]})}{P(x'_{o[i]}|\mathbf{x}^{(i)}_{-o[i]})} \quad (4)$$

The last equality follows from substituting (3), since $\mathbf{x}^{(i-1)}$ and $\mathbf{x}^{(i)}$ only differ in the $o[i]$th variable, $X_{o[i]}$.

Letting $\phi_i(\mathbf{X} = \mathbf{x}) = P(x_{o[i]}|\mathbf{x}^{(i)}_{-o[i]})/P(x'_{o[i]}|\mathbf{x}^{(i)}_{-o[i]})$ and $Z = 1/P(\mathbf{x}')$:

$$\frac{1}{Z}\prod_i \phi_i(\mathbf{x}) = P(\mathbf{x}')\frac{P(\mathbf{x})}{P(\mathbf{x}')} = P(\mathbf{x})$$

Therefore, a Markov network with the factors $\{\phi_i\}$ exactly represents the probability distribution $P(\mathbf{X})$. Since the factors are defined only in terms of conditional distributions of one variable given evidence, any consistent dependency network can be converted to a Markov network representing the exact same distribution. This holds for any ordering $o$ and base instance $\mathbf{x}'$.

Note that, unlike $\mathbf{x}'$, $\mathbf{x}$ is not a fixed vector of values but can be set to be any instance in the state space. This is necessary for $\phi_i$ to be a function $\mathbf{x}$. The following subsection will make this clearer with an example.

### 4.1 EXAMPLE

We now show how this idea can be applied to a simple, consistent DN. Consider the following conditional distributions over binary variables $X_1$ and $X_2$:

$$P(X_1 = T|X_2 = T) = 4/5$$
$$P(X_1 = T|X_2 = F) = 2/5$$
$$P(X_2 = T|X_1 = T) = 2/3$$
$$P(X_2 = T|X_1 = F) = 1/4$$

Let $\mathbf{x}' = [T, T]$ and $o = [1, 2]$. Following the earlier construction:

$$\phi_1(x_1, x_2) = \frac{P(x_1|\mathbf{x}^{(1)}_{-1})}{P(x'_1|\mathbf{x}^{(1)}_{-1})} = \frac{P(x_1|x_2)}{P(X_1 = T|x_2)}$$

$$\phi_2(x_2) = \frac{P(x_2|\mathbf{x}^{(2)}_{-2})}{P(x'_2|\mathbf{x}^{(2)}_{-2})} = \frac{P(x_2|X_1 = T)}{P(X_2 = T|X_1 = T)}$$

Note that $\phi_2$ is not a function of $x_1$. This is because $x_1^{(2)} = x'_1$, so the value of $X_1$ in the evidence is defined by the base instance, $\mathbf{x}'$. In general, the $i$th converted factor $\phi_i$ is not a function of the first $i-1$ variables, since they are fixed to values in $\mathbf{x}'$.

By simplifying the factors, we obtain the following:

| $X_1$ | $X_2$ | $\phi_1(X_1, X_2)$ |
|---|---|---|
| T | T | 1 |
| T | F | 1 |
| F | T | 1/4 |
| F | F | 3/2 |

| $X_2$ | $\phi_2(X_2)$ |
|---|---|
| T | 1 |
| F | 1/2 |

Multiplying the factors together and renormalizing yields:

| $X_1$ | $X_2$ | $P(X_1, X_2)$ |
|---|---|---|
| T | T | 0.4 |
| T | F | 0.2 |
| F | T | 0.1 |
| F | F | 0.3 |

which matches both conditional distributions. Therefore, an MN with the factors $\phi_1$ and $\phi_2$ represents the exact same distribution as a DN with the conditional distributions $P(X_1|X_2)$ and $P(X_2|X_1)$. Since the original DN is consistent, any choice of $\mathbf{x}'$ or $o$ will lead to the same result.

## 4.2 LOG-LINEAR MODELS WITH CONJUNCTIVE FEATURES

In the previous example, we showed how to convert a DN to an MN using tables as factors. However, this can be very inefficient when the conditional distributions have structure, such as decision trees or logistic regression models. Rather than treating each type of CPD separately, we discuss how to convert any distribution that can be represented as a log-linear model with conjunctive features.

Suppose the CPD for $X_i$ is a log-linear model:

$$P(X_i|\mathbf{X}_{-i}) = \frac{1}{Z(\mathbf{X}_{-i})} \exp\left(\sum_j w_j f_j(\mathbf{D}_j)\right)$$

Note that the normalization $Z$ is now a function of the evidence variables, $\mathbf{X}_{-i}$. To represent $P(X_i|\mathbf{x}_{-o[i]}^{(i)})$, we can condition each $f_j$ on $\mathbf{x}_{-o[i]}^{(i)}$ separately:

$$P(X_i|\mathbf{x}_{-o[i]}^{(i)}) = \frac{1}{Z(\mathbf{x}_{-o[i]}^{(i)})} \exp\left(\sum_j w_j f_j(X_i, \mathbf{x}_{-o[i]}^{(i)})\right)$$

$\mathbf{x}_{-o[i]}^{(i)}$ uses values from $\mathbf{x}'$ for the first $i$ variables in $o$ and $\mathbf{x}$ for the rest. The values from $\mathbf{x}'$ are constant but those in $\mathbf{x}$ are free variables, so that the resulting distribution is a function of $\mathbf{x}$. When simplifying $f_j(\mathbf{x}_{-o[i]}^{(i)})$, if the constant values in $\mathbf{x}_{-o[i]}^{(i)}$ violate one of the variable tests in $f_j$, then $f_j$ is always zero and it can be removed entirely. Otherwise, any conditions satisfied by constant values in $\mathbf{x}_{-o[i]}^{(i)}$ are always satisfied and can be removed.

Table 1: The Basic DN2MN Algorithm

**function** DN2MN($\{P_i(X_i|\mathbf{X}_{-i})\}, \mathbf{x}', o^{-1}$)
$M \leftarrow \emptyset$
**for** $i = 1$ to $n$ **do**
  Convert $P_i$ to a set of weighted features, $F_i$.
  **for** each weighted feature $(w, f) \in F_i$ **do**
    $f_n \leftarrow$ SIMPLIFYFEATURE($i, f, \mathbf{x}', o^{-1}$, true)
    $f_d \leftarrow$ SIMPLIFYFEATURE($i, f, \mathbf{x}', o^{-1}$, false)
    $M = M \cup (w, f_n) \cup (-w, f_d)$
  **end for**
**end for**
**return** $M$

For example, suppose $P(X_2|X_1, X_4)$ uses the following three conjunctive features:

$$f_1(X_1, X_2, X_4) = X_1 \wedge \neg X_2 \wedge X_4$$
$$f_2(X_1, X_2) = X_1 \wedge X_2$$
$$f_3(X_2, X_4) = X_2 \wedge \neg X_4$$

If $\mathbf{x}' = [T, T, T, T]$ and $o = [1, 2, 3, 4]$, then $\mathbf{x}^{(2)} = [X_1, X_2, T, T]$. After conditioning $f_1$, $f_2$, and $f_3$ on $\mathbf{x}_{-o[2]}^{(2)}$, they simplify to:

$$f_1(X_1, X_2) = X_1 \wedge \neg X_2$$
$$f_2(X_1, X_2) = X_1 \wedge X_2$$

$f_3$ is removed entirely, since it is inconsistent with $\mathbf{x}_{-o[2]}^{(2)}$.

To compute $\phi_2$, we must additionally condition the function in the denominator on $X_2 = \mathbf{x}'_2 = T$. If the weights of $f_1$ and $f_2$ are $w_1$ and $w_2$, respectively, then:

$$\begin{aligned}
&\phi_2(X_1, X_2) \\
&= \frac{P(X_2|X_1, X_4 = T)}{P(X_2 = T|X_1, X_4 = T)} \\
&= \frac{\exp(w_1(X_1 \wedge \neg X_2) + w_2(X_1 \wedge X_2))/Z(\mathbf{x}_{-o[2]}^{(2)})}{\exp(w_2(X_1 \wedge X_2))/Z(\mathbf{x}_{-o[2]}^{(2)})} \\
&= \exp(w_1(X_1 \wedge \neg X_2) + w_2(X_1 \wedge X_2) - w_2 X_1)
\end{aligned}$$

Note that the final factor $\phi_i$ is always a log-linear model where each feature is a subset of one of the features in the original conditional distribution. Therefore, converting each conditional distribution in this manner yields an MN represented as a log-linear model.

We summarize our complete method for consistent DNs with the algorithms in Tables 1 and 2. DN2MN takes a set of conditional probability distributions, $\{P_i(X_i|\mathbf{X}_{-i})\}$, a base instance $\mathbf{x}'$, and an inverse variable ordering $o^{-1}$. The inverse variable ordering is a mapping from variables to their corresponding indices in the desired ordering $o$, so that $o^{-1}[o[i]] = i$. DN2MN first converts the conditional

Table 2: Feature Simplifying Subroutine

**function** SIMPLIFYFEATURE($i, f, \mathbf{x}', o^{-1}$, *numerator*)
  $f' \leftarrow \emptyset$
  **for** each variable test $(X_j = v_j) \in f$ **do**
    **if** $o^{-1}[i] < o^{-1}[j]$ OR ($i = j$ AND *numerator*) **then**
      $f' \leftarrow f' \cup (X_j = v_j)$
    **else if** $v_j \neq x'_j$ **then**
      return $\emptyset$
    **end if**
  **end for**
  return $f'$

distributions into sets of weighted features. Then it conditions each feature on some of the values in $\mathbf{x}'$, depending on the variable order, to produce the simplified distributions for the numerator and denominator of (4). This is handled by SIMPLIFYFEATURE, which accepts a parameter to indicate if the simplified feature is for the numerator or denominator. Features in the denominator are assigned negative weights, since $1/\exp(wf) = \exp(-wf)$.

## 5 HANDLING INCONSISTENCIES

When the basic DN2MN algorithm is applied to an inconsistent DN, the result may depend on $\mathbf{x}'$ and $o$. For example, consider the following conditional distributions:

$$P_1(X_1 = T | X_2 = T) = 4/5$$
$$P_1(X_1 = T | X_2 = F) = 1/5$$
$$P_2(X_2 = T | X_1 = T) = 1/5$$
$$P_2(X_2 = T | X_1 = F) = 4/5$$

$P_1$ encourages $X_1$ and $X_2$ to be equal, while $P_2$ encourages them to have opposite values. By symmetry, it is easy to see that a Gibbs sampler with these transition probabilities must converge to a uniform distribution. However, the conditional distributions are not consistent with a uniform distribution, so the DN is inconsistent. We see that the choice of $\mathbf{x}'$ here does affect the converted MN:

When $\mathbf{x}' = [T, T]$ and $o = [1, 2]$:

| $X_1$ | $X_2$ | $P(X_1, X_2)$ |
|---|---|---|
| T | T | 4/85 |
| T | F | 16/85 |
| F | T | 1/85 |
| F | F | 64/85 |

When $\mathbf{x}' = [F, F]$ and $o = [1, 2]$:

| $X_1$ | $X_2$ | $P(X_1, X_2)$ |
|---|---|---|
| T | T | 64/85 |
| T | F | 1/85 |
| F | T | 16/85 |
| F | F | 4/85 |

Unsurprisingly, changing the ordering $o$ also changes the resulting probability distribution for this example. However, no single choice of $\mathbf{x}'$ and $o$ leads to the true, uniform distribution. To obtain a uniform distribution, we must average over several conversions.

### 5.1 AVERAGING OVER ALL BASE INSTANCES

A reasonable way to average is to convert the DN to an MN $k$ times and take the geometric mean of the resulting distributions. With a log-linear model, this can be done by simply combining all features from all models and dividing the weights by $k$.

However, instead of merely summing over a small number of base instances $\mathbf{x}'$, we can sum over *all* base instances in linear time and space by exploiting the structure of conjunctive features. Given a feature $f_j$ and an intermediate instance $\mathbf{x}^{(i)}_{-o[i]}$, suppose that $k$ binary-valued variables in $f_j$ have constant values in $\mathbf{x}^{(i)}_{-o[i]}$. Out of all possible $\mathbf{x}'$, only the fraction $2^{-k}$ will be consistent with the instance. Therefore, rather than enumerating all possible $\mathbf{x}'$, we can simply multiply the weight of $f_j$ by $2^{-k}$, to reflect the fact that it would be included in only that fraction of the converted MNs.

For inconsistent DNs, a uniform distribution over all $\mathbf{x}'$ may not be a good approach, since the DN may be more inconsistent in less probable regions of the state space, leading to worse behavior. We have found that a product of marginals distribution over $\mathbf{x}'$ works well. We use training data to estimate the marginal distribution of each variable, and then use those marginals to compute the modified weight of each conditioned feature. The uniform approach remains a special case. We also experimented with averaging over all training examples, but found that using the marginals worked slightly better.

### 5.2 AVERAGING OVER ORDERINGS

Since the conversion depends on the ordering, we would like to average over all possible orderings as well. Since every feature in the final model is a subset of one of the original features, a feature with $k$ variable tests has at most $2^k$ subsets to consider. When converting the conditional distribution for variable $X_i$, all variables that come after $i$ in the ordering are constant, so variable tests involving those variables will be removed. Consider a specific conjunctive subfeature that contains only $l$ of the original $k$ variable tests. There are $k!$ total orderings of the variables in the $k$ conditions. There are $(l - 1)!$ ways to order the remaining conditions (excluding the target), and $(k - l)$ ways to order the conditions that were removed.[1] Therefore, the fraction of orderings consistent with a particular subfeature is: $(l - 1)!(k - l)!/k!$.

In a model with long conjuncts, the exponential number of subfeatures leads to prohibitively large model sizes. Alternately, we can sum over a *linear* number of orderings. Given a base ordering, $o$, we sum over all $n$ orderings of the form:

$$[o[i], o[i + 1], \ldots, o[n], o[1] \ldots, o[i - 1]]$$

---
[1] We assume each variable appears in at most one condition, so they can be ordered independently. This assumption can easily be relaxed.

for all $i$. When averaging over a linear number of orderings, a conjunctive feature with $k$ variable tests will be converted to $k$ subfeatures, each determined by which of the variables comes first in the ordering.

For example, suppose the conditional distribution for $X_7$ contains the feature $X_4 \wedge X_6 \wedge X_7 \wedge X_{13}$ and we are summing over all rotations of $[1, 2, \ldots, n]$:

- If $X_7$ is first in the order, then the conditioned feature is $X_4 \wedge X_6 \wedge X_7 \wedge X_{13}$.
- If $X_5$ or $X_6$ comes first, then $X_6$ is fixed to its value in $\mathbf{x}'$, leaving $X_4 \wedge X_7 \wedge X_{13}$.
- If $X_1$ to $X_4$ or $X_{14}$ to $X_n$ comes first, then $X_4$ and $X_6$ come before $X_7$ in the ordering, leaving $X_7 \wedge X_{13}$.
- If $X_8$ to $X_{12}$ comes first, then all variables come before $X_7$, so the conditioned feature is just $X_7$.

Note that these orderings should not be weighted equally. The weights for the four subfeatures should be multiplied by $1/n$, $2/n$, $(4 + (n - 13))/n$, and $5/n$, respectively.

For symmetry, we can average over all rotations of several different orderings. In our experiments, we average over all rotations of the orderings $[1, 2, \ldots, n]$ and $[n, n-1, \ldots, 1]$. For features of length 2, this is equivalent to summing over all orderings, since all subfeatures are included and the symmetry of the opposite orderings ensures balanced weights.

# 6 EXPERIMENTS

## 6.1 DATASETS

We used 12 standard MN structure learning datasets collected and prepared by Davis and Domingos [3], omitting EachMovie since it is no longer publicly available. All variables are binary-valued. The datasets vary widely in size and number of variables. See Table 3 for a summary, and Davis and Domingos [3] for more details on their origin. Datasets are in increasing order by number of variables.

Table 3: Dataset characteristics

| Dataset | # Train | # Tune | # Test | # Vars |
|---|---|---|---|---|
| NLTCS | 16,181 | 2,157 | 3,236 | 16 |
| MSNBC | 291,326 | 38,843 | 58,265 | 17 |
| KDDCup 2000 | 180,092 | 19,907 | 34,955 | 64 |
| Plants | 17,412 | 2,321 | 3,482 | 69 |
| Audio | 15,000 | 2,000 | 3,000 | 100 |
| Jester [6] | 9,000 | 1,000 | 4,116 | 100 |
| Netflix | 15,000 | 2,000 | 3,000 | 100 |
| MSWeb | 29,441 | 3,270 | 5,000 | 294 |
| Book | 8,700 | 1,159 | 1,739 | 500 |
| WebKB | 2,803 | 558 | 838 | 839 |
| Reuters-52 | 6,532 | 1,028 | 1,540 | 889 |
| 20 Newsgroups | 11,293 | 3,764 | 3,764 | 910 |

## 6.2 METHODS

On each dataset, we learned DNs with decision tree CPDs and logistic regression CPDs. For decision tree CPDs, each probabilistic decision tree was greedily learned to maximize the conditional likelihood of the target variable given the others. To avoid overfitting, we used the structure prior of Heckerman et al. [8], $P(S) \propto \kappa^f$, where $f$ is the number of free parameters in the model and $\kappa > 0$ is a tunable parameter. Parameters were estimated using add-one smoothing. We learned logistic regression CPDs with L1 regularization using the Orthant-Wise Limited-memory Quasi-Newton (OWL-QN) method [1]. Both $\kappa$ and $\lambda$ (the L1 regularization parameter) were tuned using held-out validation data. For $\kappa$, we started from a value of $10^{-4}$ and increased it by a factor of 10 until the score on the validation set decreased. For $\lambda$, we used the values $0.1, 0.2, 0.5, 1, 2, 5, \ldots, 1000$ and selected the model with the highest score on the validation set. To score a DN on a validation set, we computed the product of the conditional probabilities of each variable according to their respective CPDs.

We converted the DNs to MNs with 6 different choices of base instances and orderings:

1. One ordering; one base instance
2. Two opposite orderings; one base instance
3. One ordering; expectation over all instances
4. Two opposite orderings; expectation over all instances
5. All rotations of one ordering; expectation over all instances
6. All rotations of two opposite orderings; expectation over all instances

The expectation over all instances was done using the marginal distribution of the variables, as estimated using the training data. To be as fair as possible, we include the time to read in the entire training data in our timing results, as well as the time to write the model to disk.

We also converted the DNs to MNs using weight learning, similar to the methods of [15, 12]. We first converted all DN CPDs to conjunctive features and removed duplicate features. Then we learned weights for all features to maximize the PLL of the training data, using a Gaussian prior on the weights to reduce overfitting. The standard deviation of the Gaussians was tuned to maximize PLL on the validation set. With decision tree CPDs, we used standard deviations of $0.05, 0.1, 0.2, 0.5,$ and $1.0$, and the best-performing models always had standard deviations between $0.1$ and $0.5$. For logistic regression CPDs, a slightly wider range of standard deviations was required, ranging from $0.1$ to $10$. Optimization was done using the limited memory BFGS algorithm, a quasi-Newton method for convex optimization [11]. Weight learning was terminated after 100 iterations if it had not yet converged.

We did not compare to BLM [3] or the algorithm of Della Pietra et al. [4], since those algorithms have been shown to be less accurate and many orders of magnitude slower than Ravikumar et al. [15] on these datasets [12, 7].

In addition to PLL, we computed conditional marginal log-likelihood (CMLL), a common evaluation metric for Markov networks [3, 10, 12]. CMLL is the sum of the conditional log-likelihood of each variable, given a subset of the others as evidence. In our experiments, we followed the approach of Davis and Domingos [3], who partition the variables into four sets, $\mathbf{Q}_1$ through $\mathbf{Q}_4$. The marginal distribution of each variable is computed using the variables from the other three sets as evidence:

$$\text{CMLL}(\mathbf{X}) = \sum_j \sum_{X_i \in \mathbf{Q}_j} \log P(X_i | \mathbf{X} - \mathbf{Q}_j)$$

Marginals were computed using Gibbs sampling. We used a Rao-Blackwellized Gibbs sampler that adds counts to all states of a variable based on its conditional distribution before selecting the next value for that variable. We found that 100 burn-in and 1000 sampling iterations were sufficient for Gibbs sampling, and additional iterations did not affect the results very much.

All of our learning and inference methods are available in the latest version of the open-source Libra toolkit.[2]

### 6.3 RESULTS

First, we compare the different approaches to converting inconsistent DNs and compare the accuracy of the resulting MNs to the original DNs. Figure 1 shows the result of converting DNs with both decision tree CPDs (left) and logistic regression CPDs (right). We measure the relative accuracy by comparing the PLL of each MN to the original DN on the test data. In order to better show differences among datasets with different numbers of variables, we show the normalized PLL (NPLL), which divides the PLL by the number of variables in the domain. All graphs are rescaled so that the height of the tallest bar in each cluster is one, with its actual height listed above the graph. With consistent DNs, all methods would be equivalent and all differences would be zero. Since the DNs are inconsistent, converting to an MN may result in an altered distribution that is less accurate on test data.

Empirically, we see that using many orders and summing over all base instances leads to more accurate models. For the decision tree DNs, some of the bars are negative, indicating MNs that are more accurate than their source DNs. This can happen because the conditional distributions in the MN combine several conditional distributions from the original DN, leading to more complex dependencies than a single decision tree. With $2n$ orderings, the converted MN was more accurate on 10 out of 12 datasets. For logistic regression DNs, the converted MNs are very, very close in accuracy, but never better. With $2n$ orderings, the difference in NPLL is less than 0.005 for all datasets, and is less than 0.001 for 8 out of 12.

---

[2] http://libra.cs.uoregon.edu/

Table 4 shows the relative accuracy of converting DNs to MNs by learning weights (LW) or by direct conversion with $2n$ orderings and a marginal distribution over base instances (DN2MN). The result of the better performing algorithm is marked in bold. Differences on individual datasets are not always statistically significant, but the overall ranking trend shows the relative performance of the two methods. LW and DN2MN do similarly well on tree CPDs: neither PLL nor CMLL was significantly different according to a Wilcoxon signed ranks test. With logistic regression CPDs, DN2MN achieves higher PLL on 10 datasets and CMLL on 9. Both differences are significant at $p < 0.05$ under a two-tailed Wilcoxon signed ranks test.

Table 5 shows the time for converting DNs using our closed-form solution and weight learning. DN2MN is 7 to 250 times faster than weight learning with decision tree CPDs, and 160 to 1200 times faster than weight learning with logistic regression CPDs. These results include tuning time, which is necessary to obtain the best results for weight learning. If all tuning time is excluded, weight learning is 12 times slower with tree CPDs and 49 times slower with logistic regression CPDs, on average.

## 7 CONCLUSION

DN2MN learns an MN with very similar performance to the original DN, and does so very quickly. With decision tree CPDs, the MN is often more accurate than the original DN. With logistic regression CPDs, the MN is significantly more accurate than performing weight learning as done by Ravikumar et al. [15]. This makes DN2MN competitive with or better than state-of-the-art methods for learning MN structure. Furthermore, DN2MN can exploit many types of conditional distributions, including boosted trees and rule sets, and can easily take advantage of any algorithmic improvements in learning conditional distributions. Another important application of DN2MN is when the model is specified by experts, since conditional distributions could be much easier to specify than a joint distribution.

**Acknowledgements**

We thank Chloé Kiddon, Jesse Davis, and the anonymous reviewers for helpful comments. This research was partly funded by ARO grant W911NF-08-1-0242, NSF grant IIS-1118050, and NSF grant OCI-0960354. The views and conclusions contained in this document are those of the author and should not be interpreted as necessarily representing the official policies, either expressed or implied, of ARO, NSF, or the U.S. Government.

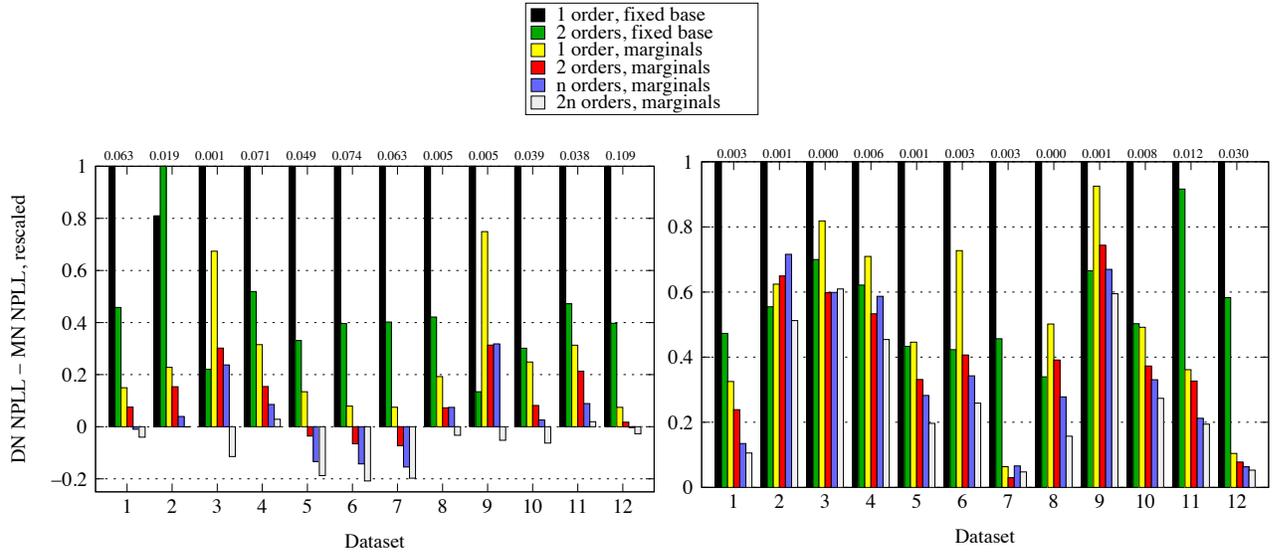

Figure 1: Difference in normalized PLL between the original DNs and converted MNs with different orderings and base instances, divided by largest difference. Results from decision tree DNs are on the left; logistic regression DNs are on the right. Smaller values indicate better MN performance. Largest difference values are listed above each dataset's results, rounded to nearest 1/1000th.

Table 4: Test set PLL and CMLL of converted DNs with tree and logistic regression CPDs.

|  | Tree CPDs | | | | LR CPDs | | | |
| --- | --- | --- | --- | --- | --- | --- | --- | --- |
|  | PLL | | CMLL | | PLL | | CMLL | |
| Dataset | LW | DN2MN | LW | DN2MN | LW | DN2MN | LW | DN2MN |
| NLTCS | -5.02 | **-4.93** | -5.25 | **-5.20** | -4.96 | **-4.95** | -5.23 | **-5.23** |
| MSNBC | -4.32 | **-4.31** | **-5.75** | -5.80 | -6.06 | **-6.06** | **-6.28** | -6.28 |
| KDDCup 2000 | **-2.05** | -2.05 | -2.08 | **-2.07** | **-2.06** | -2.07 | **-2.11** | -2.12 |
| Plants | **-8.75** | -9.17 | **-10.00** | -10.67 | **-9.39** | -9.50 | **-10.76** | -10.91 |
| Audio | -38.01 | **-37.77** | **-38.25** | -38.35 | -36.17 | **-36.11** | -36.93 | **-36.88** |
| Jester | -51.42 | **-50.77** | -51.49 | **-51.42** | -49.01 | **-48.81** | -49.83 | **-49.76** |
| Netflix | -54.32 | **-53.60** | -54.62 | **-54.50** | -51.15 | **-51.10** | -52.37 | **-52.31** |
| MSWeb | **-8.20** | -8.33 | **-8.72** | -8.77 | -8.70 | **-8.64** | -8.96 | **-8.93** |
| Book | **-34.60** | -35.14 | **-34.49** | -35.44 | -33.86 | **-33.41** | -34.75 | **-34.09** |
| WebKB | -149.37 | **-148.42** | **-149.99** | -151.88 | -153.13 | **-139.39** | -158.51 | **-143.05** |
| Reuters-52 | **-82.57** | -82.67 | **-82.53** | -85.16 | -81.44 | **-77.62** | -81.82 | **-79.60** |
| 20 Newsgroups | -159.14 | **-152.84** | -156.08 | **-154.06** | -151.53 | **-147.76** | -151.93 | **-148.82** |

Table 5: Total running time for learning MNs from DNs. DN2MN method uses $2n$ orderings and marginals.

|  | Tree CPDs | | | LR CPDs | | |
| --- | --- | --- | --- | --- | --- | --- |
| Dataset | LW | DN2MN | Speedup | LW | DN2MN | Speedup |
| NLTCS | 4.9s | 0.3s | 17.3 | 2.3s | 0.01s | 162.4 |
| MSNBC | 73.3s | 9.4s | 7.8 | 7.6s | 0.04s | 189.6 |
| KDDCup 2000 | 121.3s | 3.8s | 31.5 | 32.0s | 0.12s | 251.7 |
| Plants | 67.9s | 1.6s | 42.7 | 71.4s | 0.17s | 427.6 |
| Audio | 147.4s | 1.8s | 80.3 | 139.3s | 0.40s | 344.0 |
| Jester | 56.6s | 1.2s | 46.6 | 311.7s | 0.32s | 974.1 |
| Netflix | 109.8s | 2.0s | 54.7 | 534.3s | 0.43s | 1245.6 |
| MSWeb | 452.8s | 16.0s | 28.3 | 308.6s | 0.65s | 472.0 |
| Book | 361.6s | 1.4s | 252.6 | 224.8s | 1.10s | 203.5 |
| WebKB | 177.5s | 1.8s | 128.6 | 386.3s | 1.71s | 225.9 |
| Reuters-52 | 798.2s | 3.3s | 242.7 | 951.0s | 2.66s | 357.7 |
| 20 Newsgroups | 1952.9s | 7.6s | 257.9 | 3830.5s | 7.61s | 503.0 |
| Geom. mean | 149.3s | 2.5s | **59.9** | 145.4s | 0.38s | **363.9** |